\documentclass[11.5pt]{article}

\usepackage[utf8]{inputenc}
\usepackage{natbib}
\usepackage{fullpage}

\usepackage{scalerel} 
\usepackage{bm}
\usepackage{bbm}
\usepackage{mathrsfs}
\usepackage{mathtools}
\usepackage{dsfont}
\usepackage{booktabs}
\usepackage{amsfonts, amsmath, amssymb, amsthm}

\usepackage{changepage}
\usepackage{adjustbox}
\usepackage{graphicx}
\usepackage{pgfplots}
\pgfplotsset{compat=newest}
\usetikzlibrary{plotmarks}
\usetikzlibrary{arrows.meta}
\usepgfplotslibrary{patchplots}
\usepackage{grffile}

\usepackage[ruled]{algorithm2e}

\newtheorem{theorem}{Theorem}

\newtheorem{proposition}{Proposition}
\newtheorem{assu}{Assumption}

\usepackage{xcolor}

\definecolor{DarkGreen}{rgb}{0.1,0.6,0.1}
\definecolor{DarkRed}{rgb}{0.6,0.1,0.1}
\definecolor{DarkBlue}{rgb}{0.1,0.1,0.6}

\usepackage[breaklinks,colorlinks,bookmarks=true]{hyperref}
\hypersetup{
    citecolor=DarkBlue,  
    linkcolor=DarkRed
}

\usepackage[capitalize,nameinlink,noabbrev]{cleveref}
\crefname{section}{\S}{\S\S}
\Crefname{section}{\S}{\S\S}

\newcommand{\ci}{\mathrel{\perp\!\!\!\perp}}
\newcommand{\Span}{\mathrm{span}}
\newcommand{\E}{\mathbb{E}}
\newcommand{\Var}{\mathrm{Var}}
\newcommand{\Cov}{\mathrm{Cov}}
\newcommand{\Corr}{\mathrm{Corr}}
\newcommand{\R}{\mathbb{R}}

\newcommand{\X}{\mathcal{X}}

\newcommand{\Yhat}{\vstretch{0.9}{\widehat{Y}}}

\newcommand{\one}{\mathds{1}}
\newcommand{\binset}{\{0, 1\}}

\title{Learning Fair Representations for Kernel Models}
\author{Zilong Tan\\ Carnegie Mellon University\\ zilongt@cs.cmu.edu
\and Samuel Yeom\\ Carnegie Mellon University\\ syeom@cs.cmu.edu
\and Matt Fredrikson\\ Carnegie Mellon University\\ mfredrik@cs.cmu.edu
\and Ameet Talwalkar\\ Carnegie Mellon University\\ \& Determined AI\\ talwalkar@cmu.edu}
\date{}

\begin{document}

\maketitle

\begin{abstract}
Fair representations are a powerful tool for satisfying fairness goals such as statistical parity and equality of opportunity in learned models. 
Existing techniques for learning these representations are typically \emph{model-agnostic}, as they pre-process the original data such that the output satisfies some fairness criterion, and can be used with arbitrary learning methods.
In contrast, we demonstrate the promise of learning a \emph{model-aware} fair representation, focusing on kernel-based models.
We leverage the classical sufficient dimension reduction (SDR) framework to construct representations as subspaces of the reproducing kernel Hilbert space (RKHS), whose member functions are guaranteed to satisfy a given fairness criterion.
Our method supports several fairness criteria, continuous and discrete data, and multiple protected attributes.
We also characterize the fairness-accuracy trade-off with a parameter that relates to the principal angles between subspaces of the RKHS.
Finally, we apply our approach to obtain the first fair Gaussian process (FGP) prior for fair Bayesian learning, and show that it is competitive with, and in some cases outperforms, state-of-the-art methods on real data.
\end{abstract}

\section{Introduction}

Fairness has emerged as a key issue in machine learning as the learned models are increasingly used in areas such as hiring~\citep{dastin2018amazon}, healthcare~\citep{gupta2017advances}, and criminal justice~\citep{compas}.
In particular, the models' predictions should not lead to decisions that discriminate on the basis of a legally protected attribute, such as race or gender.
Among the proposals to address this issue, a growing body of work focuses on learning fair representations of data for downstream modeling~\citep{calmon2017optimized, delbarrio2018obtaining, feldman2015certifying, johndrow2019algorithm, kamiran2012data}.
Most of these approaches are \emph{model-agnostic}, which provides flexibility when working with the learned representations but comes at the cost of potentially suboptimal results in terms of both fairness and accuracy.

In this work, we present a novel approach for fair representation learning that takes into account the target hypothesis space of models that will be learned from the representation.
Specifically, we show how to leverage information about the reproducing kernel Hilbert space (RKHS) to learn a fair representation for kernel-based models with provable fairness and accuracy guarantees.

Our approach builds on the sufficient dimension reduction (SDR) framework \citep{Li91,Fukumizu09,Wu09}, which is used to compute a low-dimensional projection of the feature vector $X$ that captures all information related to the response $Y$. Our key insight is that we can instead perform SDR with respect to the protected attributes $S$, and then take the orthogonal complement of the resulting projection to obtain a {\em fair subspace} of the RKHS that captures information in $X$ unrelated to $S$. We show that functions in the fair subspace will be independent of $S$ under mild conditions (\cref{sec:fair-rep}), and we leverage this fact to prove that our approach can guarantee several popular definitions of fairness, namely statistical parity~\citep{feldman2015certifying}, proxy nondiscrimination~\citep{datta2017proxy}, equality of opportunity~\citep{hardt2016equality}, and equalized odds~\citep{hardt2016equality}. Moreover, our approach is compatible with both classification and regression, as well as in settings where there are multiple, possibly continuous protected attributes.

Because a fair model might have a lower-than-desired accuracy in practice, we further generalize our approach to consider this trade-off. In particular, we apply SDR to compute a \emph{predictive subspace} of the RKHS that captures sufficient information in the feature vector $X$ related to the response $Y$.  We then define a third {\em model subspace} of the RKHS, which is bounded between the fair and predictive subspaces by a specified principal angle~\citep{Stewart90,Golub13}. In contrast to recent regularization- and constraint-based trade-offs \citep{zemel2013learning,madras2018learning,louizos2015variational,Edwards16,Song19}, we provide precise characterizations of how the specified angle affects the fairness and accuracy of any model in this subspace.

Finally, we apply our method to obtain, to the best of our knowledge, the first fair Gaussian process (FGP) prior for constructing fair models in the Bayesian setting. Sample paths 
of the FGP will be functions in the chosen model subspace, and hence satisfy the specified fairness conditions. 
We identify the covariance kernel of the FGP that corresponds to the chosen model subspace by using the duality between a Gaussian process and its RKHS \citep{Wahba90,Pillai07,Tan18}. Our experiments 
show that the FGP achieves both rigorous fairness properties and improved accuracy compared to prior methods.

\paragraph{Outline}
The remainder of the paper is organized as follows. In Section~\ref{sec:fairness}, we introduce an SDR-induced fair representation and show that it satisfies several existing fairness criteria. In Section~\ref{sec:alg}, we present an algorithm for computing the proposed fair representation that attains a specified fairness-accuracy trade-off. We provide theoretical analysis that characterizes this trade-off as well as generalization performance. In Section~\ref{sec:fairgp}, we demonstrate the application of the fair representation to GPs, and Section~\ref{sec:exp} reports empirical results of the fair GP on several real-life datasets.

\paragraph{Additional Related Work}

Much of the prior work on fair representation learning optimizes only for statistical parity~\citep{zemel2013learning,delbarrio2018obtaining,feldman2015certifying,johndrow2019algorithm,louizos2015variational,komiyama2017two,komiyama2018nonconvex} or individual fairness~\citep{calmon2017optimized}. Learned Adversarially Fair and Transferable Representations (LAFTR)~\citep{madras2018learning} provides additional support for equality of opportunity and equalized odds by taking into account the model loss while learning the fair representation. The authors prove bounds for statistical parity and equalized odds, but it should be noted that these bounds depend on the optimal adversary, which may not be available in non-convex settings. Our approach supports a broader set of fairness criteria (see \cref{sec:indep}), and we characterize the generalization performance in terms of both fairness and accuracy.

Provably fair kernel learning has been recently studied by~\cite{Donini18} and~\cite{Oneto19}. Both approaches primarily target equality of opportunity in the setting of a single protected attribute. As previously noted, we address a more general setting. \cite{komiyama2018nonconvex} also study fair kernel methods, but they only remove linear correlation between the input features $X$ and the protected attribute $S$; by contrast, we can remove general statistical associations between $X$ and $S$.

Bayesian formulations of fairness have been studied by \citet{Foulds18} and \citet{Dimitrakakis19}, who take into account the uncertainty in model parameters. \cite{Dimitrakakis19} propose Bayesian versions of the balance \citep{kleinberg2017inherent} and calibration criteria \citep{chouldechova2017fair} based on decision-theoretic risk formulations. \cite{Foulds18} introduce a differential fairness criterion inspired by the definition of differential privacy in the setting of multiple protected attributes. Unlike these works, we do not propose new fairness criteria, instead focusing on several widely used criteria as described in \cref{sec:indep}.

Much previous research on learning rules with explicit fairness constraints or objectives~\citep{Kamiran10,Zliobaite15,Zafar17a} includes empirical studies on the fairness-accuracy trade-off, reporting that classifiers trained in this way outperform those trained on model-agnostic fair representations. Our proposed fair representations are not model-agnostic, and our performance is competitive if not better in some cases than that of those learning methods. \cite{Menon18} provide a theoretical analysis of the trade-off, providing information-theoretic bounds on accuracy in terms of the correlation between the target and protected attributes, as well as a regularization parameter analogous to the principal angle between subspaces used to set the trade-off in our work. In contrast, we provide insight into how the trade-off impacts generalization performance. 

Another approach to fair classification uses randomized post-processing of the classifier's predictions to ensure group fairness criteria. \cite{hardt2016equality} propose such a procedure for ensuring equalized odds on binary classifiers. \cite{Woodworth17} argue that this approach can be suboptimal, and propose an alternative scheme: first learn a classifier with constraints to approximate fairness, and subsequently post-process its predictions to reduce discrimination. These approaches are orthogonal to fair representation learning, and do not consider either regression or multiple protected attributes.

\section{Using SDR to Formulate Fairness}
\label{sec:fairness}

We begin by introducing some notation. We write $\X$ for the feature space and $\mathcal{S}$ for the space of protected attributes. In addition, $X \in \X$, $S \in \mathcal{S}$, and $Y \in \R$ denote the random variables for the feature vector, protected attributes, and label/response, respectively. We write accordingly $\left\{\left(\bm{x}_i,s_i,y_i\right) \in \mathcal{X} \times \mathcal{S} \times \R\right\}_{i=1}^n$ for the training data of $n$ examples.

The fair subspace is chosen as a vector space such that the projection of $X$ onto the fair subspace does not contain information about $S$ while retaining the residual information of $X$. We then use the projection as the fair representation. In the next subsection, we present an SDR-based method, i.e., \cref{prop:SDR-fairness}, for finding a fair subspace. In \cref{sec:indep}, we show that the fair subspace based representation satisfies several existing fairness criteria.

\subsection{SDR-Induced Fair Representations}
\label{sec:fair-rep}

First consider the simple case where $\X = \R^p$ and $\mathcal{S} = \R$, including both the categorical and continuous cases. The goal is to obtain a basis of the fair subspace of dimension $d \leq p$ represented by the columns of $\bm{C} \in \R^{p\times d}$. A salient challenge in finding the fair subspace arises as the link between $S$ and $X$ is unknown. To address this issue, the key insight we use is that $\bm{C}$ can be obtained from an {\em SDR subspace} of $X$ with respect to $S$. Next, we briefly recall the definition of the SDR subspace as well as its assumption \citep{Li91,Cook09}. Then, we provide the construction of $\bm{C}$ in \cref{prop:SDR-fairness}.

An $m$-dimensional vector space is called an \emph{SDR subspace} of $X$ with respect to $S$ if the projection of $X$ onto the subspace captures the statistical dependency of $S$ on $X$. The proposed SDR-induced fair representation relies on a kernel version of the following SDR assumption.
\begin{assu}
\label{assu:SDR}
There exists a function $f_S: \R^m\times \R^l \to \R$ and a matrix $\bm{B} \in \R^{p \times m}$ such that
\begin{align}
\label{eq:SDR-S}
    S = f_S\left(\bm{B}^\top X,\epsilon_S\right) \quad \text{or} \quad
    X \ci S \mid \bm{B}^\top X,
\end{align}
where $\epsilon_S$ is a random variable independent of $X$.
\end{assu}
The column span of $\bm{B}$ is known as the SDR subspace. It is worth pointing out that the condition \eqref{eq:SDR-S} is {\em always} satisfied since there is a one-to-one correspondence between $\bm{B}^\top X$ and $X$ whenever $\bm{B}$ is square non-singular. The condition \eqref{eq:SDR-S} states that the projection $\bm{B}^\top X$ captures all information in $X$ about $S$ or more. The goal is thus to recover an SDR subspace with the lowest dimension. Under mild conditions on $X$, this recovery is guaranteed without requiring the knowledge of $f_S$~\citep{Li91,Hall93}.

The SDR-induced fair representation is given by the projection onto the fair subspace $X^\prime \coloneqq \bm{C}^\top X$, where $\bm{C}$ is defined as in \cref{prop:SDR-fairness}. Note that the proposition does not make assumptions about the underlying link function $f_S$. Its proof adapts techniques used in \citealp{Brillinger83} as well as the properties of elliptically contoured distributions, and is provided in \cref{sec:proofs}. 

\begin{proposition}
\label{prop:SDR-fairness}
Suppose that $\E\left|S\right| < \infty$ and $\E \left|X_i  S\right| < \infty$ for $i=1,\cdots,p$. Let the columns of $\bm{C}$ form a basis of the nullspace of $\Var\left(X\right) \bm{B}$. If condition \eqref{eq:SDR-S} holds and $X$ follows an elliptically contoured distribution, then $\Cov \left(\bm{C}^\top X,S\right) = \bm{0}$.
\end{proposition}

The class of elliptically contoured distributions contains the normal distribution. In the case where $X^\prime$ and $S$ are jointly multivariate normal, the lack of correlation guaranteed by \cref{prop:SDR-fairness} implies $X' \ci S$. We remark that the multivariate normal requirement of the pair $\left(X^\prime,S\right)$ is reasonable for high-dimensional $X$, as most low-dimensional projections of high-dimensional data are nearly normal under mild conditions~\citep{Diaconis84,Hall93}. Moreover, the high-dimensional condition holds for kernel models where input data is mapped to potentially infinite-dimensional feature space.

\paragraph{Extensions to Multivariate $S$ and RKHS}
Our approach incorporates two generalizations of the linear SDR condition \eqref{eq:SDR-S}. First, the condition \eqref{eq:SDR-S} does not apply to the setting with multiple protected attributes. This is handled by replacing \eqref{eq:SDR-S} with the joint conditions $S_i = f_{S_i}\left(\bm{B}^\top X,\epsilon_{S_i}\right)$ for each protected attribute $S_i$ \citep{Coudret14}. Second, $S$ can depend on nonlinear structures of $X$, and hence the linear condition \eqref{eq:SDR-S} may not yield a low-rank $\bm{B}$. In that case, the resulting fair subspace $\bm{C}^\top X$ may be too low-dimensional for accurately predicting $Y$. To address this issue, we use the RKHS counterpart of condition \eqref{eq:SDR-S}. 

Denote by $\mathcal{H}_\kappa$ an RKHS of functions $f: \mathcal{X} \mapsto \R$ generated by kernel $\kappa\left(\cdot,\cdot\right)$. In the RKHS setting with training points $\bm{x}_1, \ldots, \bm{x}_n$, $X$ is replaced by the feature function $\kappa\left(\cdot,X\right)$, and $\bm{B}_i$ will instead be functions in $\mathcal{H}_\kappa$ expressed as $\bm{B}_i = \sum_{j=1}^n W_{ji} \kappa\left(\cdot,\bm{x}_j\right)$ for some $\bm{W} \in \R^{n\times m}$ by the representer theorem \citep{Scholkopf01}. Thus, condition \eqref{eq:SDR-S} in the RKHS setting reads
\begin{align}
\label{eq:SDR-RKHS}
    S = f_S\left(\kappa\left(X,\bm{X}\right)\bm{W}_1,\cdots,\kappa\left(X,\bm{X}\right)\bm{W}_m, \epsilon_S\right)
\end{align}
with $\kappa\left(X,\bm{X}\right) \coloneqq \left(\kappa\left(X,\bm{x}_1\right),\dots,\kappa\left(X,\bm{x}_n\right)\right)$. Similarly, we also adapt \cref{prop:SDR-fairness} to the RKHS setting with $\bm{C}_i$ replaced by $\sum_{j=1}^n Q_{ji}\kappa\left(\cdot,\bm{x}_j\right)$ for some $\bm{Q} \in \R^{n\times r}$. This yields the corresponding fair subspace \begin{align}
\label{eq:fair-subspace}
    \Span\left\{\sum_{j=1}^n Q_{j1}\kappa\left(\cdot,\bm{x}_j\right),\dots,\sum_{j=1}^n Q_{jr}\kappa\left(\cdot,\bm{x}_j\right)\right\},
\end{align}
and the fair representation $X^\prime = \kappa\left(X,\bm{X}\right)\bm{Q}$ of $X$.

\paragraph{Kernel Misspecification}
In the RKHS setting, \cref{prop:SDR-fairness} states that the fair representation $X^\prime$ does not contain information about $S$. However, this does not necessarily imply that $X^\prime$ is predictive in terms of $Y$. For example, one can trivially satisfy \eqref{eq:SDR-RKHS} by letting $\bm{W}$ be the identity matrix. Then, the corresponding fair representation  will be the constant $0$, which is trivially independent of $X$. Thus, in order for the fair representation to be predictive, $\kappa\left(\cdot,\cdot\right)$ and $\bm{W}$ should be chosen such that $\eqref{eq:SDR-RKHS}$ holds for a small $m$.

We note that a misspecified kernel affects the predictive power, but not fairness, of the fair representation as long as \eqref{eq:SDR-RKHS} holds. Kernel misspecification can be detected in practice because the resulting model based on $X^\prime$ would give a large prediction error. The problem of finding appropriate $\bm{W}$ and $m$ will be discussed in \cref{sec:learn-basis}.

\subsection{Fairness as Statistical Independence}
\label{sec:indep}

In this section, we formulate several common fairness criteria in terms of the statistical independence $X^\prime \ci S$, where $X^\prime$ is the projection of $X$ onto the fair subspace described in \cref{sec:fair-rep}. Let $h\left(\cdot\right)$ denote the model and let $\Yhat = h(X^\prime)$ be the model output. This paper focuses on the following fairness criteria:

\begin{itemize}
\item \emph{Statistical parity} (SP), also called \emph{demographic parity}, is one of the simplest notions of fairness and requires model predictions to be independent of the protected attributes, i.e., $\Yhat \ci S$. This follows from $X' \ci S$ since $\Yhat$ is a function of $X'$.
\item \emph{Proxy nondiscrimination}~\citep{datta2017proxy, yeom2018hunting} goes further than statistical parity in that it considers all components of the model rather than just its output. For example, a component $\bm{c}$ of a linear model $h\left(X\right) = \bm{\beta}^\top X$ has the output $\Yhat_{\bm{c}} \coloneqq \sum_{i=1}^p c_i \beta_i X_i$ for $c_i \in \left[0,1\right]$.
The strictest version of proxy nondiscrimination requires $\Yhat_{\bm{c}} \ci S$ for all component $\bm{c}$.
This follows from $X' \ci S$ since $\Yhat_{\bm{c}}$ is a function of $X'$.
\item {\em {Equalized Odds, Equality of Opportunity}:} In the binary classification setting where $Y \in \binset$, the equalized odds (EO) condition \citep{hardt2016equality} is defined as the conditional independence $\Yhat \ci S \mid Y$. Compared to statistical parity, one advantage of equalized odds is that it admits the perfect model $\Yhat = Y$. Equality of opportunity (EOP) \citep{hardt2016equality} is a relaxation of equalized odds, requiring only that $\Yhat \ci S \mid Y{=}1$. To attain EOP, we can apply \eqref{eq:SDR-S} to only the individuals with $Y{=}1$, and use the resulting $X^\prime$ as input features. Similarly, we can achieve EO by restricting \eqref{eq:SDR-S} to individuals with $Y{=}1$ to obtain $\bm{B}_{Y{=}1}$ and to individuals with $Y{=}0$ to obtain $\bm{B}_{Y=0}$. Then, we compute $X^\prime$ by taking the union of SDR subspaces $\left[\bm{B}_{Y{=}1} \enspace \bm{B}_{Y{=}0}\right]$ as the $\bm{B}$ in \cref{prop:SDR-fairness}.
\end{itemize}

We conclude the discussion by pointing out that our approach does not support accuracy parity \citep{Zafar17}, which requires $\one(\Yhat = Y) \ci S$, or the calibration condition $Y \ci S \mid \Yhat$ \citep{chouldechova2017fair}. This is because, without further assumptions on the model, a fair representation alone cannot preclude a constant model. If $\Yhat$ is a constant, to satisfy accuracy parity and calibration we would need some independence condition between $Y$ and $S$, which does not in general hold. Thus, it could be interesting future work to further identify additional conditions on the model needed to support the other fairness definitions.

\section{Computing the Hypothesis Space}
\label{sec:alg}

We now describe how to compute the fair subspace as well as the model subspace in which every function attains a desired fairness-accuracy trade-off. The construction is presented analytically in \eqref{eq:model-basis}, and we provide generalization bounds in \cref{thm:proj-pert} and \eqref{eq:generalization} for the deviation between an optimal fair or predictive model in the RKHS and the model obtained from the model subspace on unseen data. The pseudo-code and all proofs are provided in the appendix.

\paragraph{Problem Setup}
Recall that a kernel-based learning problem is typically formulated as the following optimization under Tikhonov regularization \citep{Cucker02,Hofmann08}:
\begin{align}
\label{eq:reg-obj}
\min_{f \in \mathcal{H}_{\kappa}} L\left(f,\left\{\left(\bm{x}_i,y_i\right)\right\}_{i=1}^n\right) + R\left(\left\|f\right\|_{\mathcal{H}_{\kappa}}\right),
\end{align}
where $L$ is a convex loss, $R$ is a monotonically increasing regularization function, and $\{(\bm{x}_i, y_i)\}_{i=1}^n$ is the set of training points. While $\mathcal{H}_\kappa$ is infinite dimensional, the well-known representer theorem \citep{Wahba90,Scholkopf01} states that the solution $f_\star$ for \eqref{eq:reg-obj} is in a data-dependent finite-dimensional subspace of $\mathcal{H}_\kappa$:
\begin{align}
\label{eq:finite-universe}
    \mathcal{H}_{\kappa,n} \coloneqq \left\{\sum_{i=1}^n a_i \kappa\left(\cdot,\bm{x}_i\right) \;\Big|\; \left\{a_i\right\}_{i=1}^n \subset \R \right\}.
\end{align}

Our goal is to obtain a subset of functions in $\mathcal{H}_{\kappa,n}$ that meets the fairness criteria described in \cref{sec:indep}. This subset will be the fair subspace \eqref{eq:fair-subspace} of $\mathcal{H}_{\kappa,n}$.

\subsection{Learning the Fair and Predictive Subspaces}
\label{sec:learn-basis}

Both the predictive subspace $\mathcal{G}$ and the fair subspace $\mathcal{F}$ of $\mathcal{H}_\kappa$ can be estimated using a standard SDR estimator for the RKHS. Specifically, the predictive subspace $\mathcal{G}$ is given by the SDR subspace with respect to $Y$. Similarly, the fair subspace $\mathcal{F}$ given by \eqref{eq:fair-subspace} is obtained by first computing the SDR subspace with respect to $S$ and then using \cref{prop:SDR-fairness} to obtain $\bm{Q}$.


Since $\mathcal{G}$ and $\mathcal{F}$ are subspaces of $\mathcal{H}_{\kappa,n}$, we write $\mathcal{G} = \Span\left\{\bm{\phi}\bm{A}_1,\cdots,\bm{\phi}\bm{A}_d\right\}$ and $\mathcal{F} = \Span\left\{\bm{\phi}\bm{Q}_1,\cdots,\bm{\phi}\bm{Q}_r\right\}$  with $r = n - m$ and $\bm{\phi} \coloneqq \left(\kappa\left(\cdot,\bm{x}_1\right),\cdots,\kappa\left(\cdot,\bm{x}_n\right)\right)$.  Our goal is thus to compute $\bm{A}$ and $\bm{Q}$, the latter of which requires the SDR subspace specified by $\bm{W}$. Next we briefly review the estimation of $\bm{A}$; $\bm{W}$ is obtained similarly with respect to $S$. In the following, we assume without loss of generality that $d + m \leq n$.

\paragraph{Estimating the SDR Subspace}
The estimate of $\bm{A}$ is given by the eigenvectors of the following generalized eigenvalue decomposition \citep{Tan18}:
\begin{align}
\label{eq:rkhs-eig}
\bm{\Gamma}_n\bm{K} \bm{A}_i = \tau_i \left(\bm{\Delta} + n \eta\bm{I}_n\right) \bm{A}_i,
\end{align}
where $\bm{\Gamma}_n \coloneqq \bm{I}_n - \bm{1}_n\bm{1}_n^\top/n$, $\bm{K}$ represents the kernel matrix with $K_{ij} \coloneqq \kappa\left(\bm{x}_i,\bm{x}_j\right)$, $\eta > 0$ is a regularization parameter, and $\bm{\Delta}$ is a matrix to be discussed shortly. For simplicity, \eqref{eq:rkhs-eig} assumes that the data tuples $\left(\bm{x}_i,\bm{s}_i,y_i\right)$ are sorted by $y_i$, either in ascending or descending order. To obtain $\bm{\Delta}$, one first partitions the data into slices $\left\{\left(\bm{x}_1,\bm{s}_1,y_1\right),\cdots,\left(\bm{x}_{n_1},\bm{s}_{n_1},y_{n_1}\right)\right\}$, $\left\{\left(\bm{x}_{n_1 + 1},\bm{s}_{n_1+1},y_{n_1+1}\right),\cdots,\left(\bm{x}_{n_1+n_2},\bm{s}_{n_1+n_2},y_{n_1+n_2}\right)\right\}$, and so forth, where $n_i$ denotes the size of the $i$-th slice. Then, set $\bm{\Delta} = \text{diag}\left(\bm{\Gamma}_{n_i}\right)\bm{K}$ where $\text{diag}\left(\bm{\Gamma}_{n_i}\right)$ is the block-diagonal matrix with diagonal blocks $\bm{\Gamma}_{n_i}$. The overall computational complexity for estimating $\bm{A}$ is $O\left(n^2 d\right)$.

Another relevant problem is to decide the dimensions of $\mathcal{G}$ and $\mathcal{F}$, i.e., the values of $m$ and $d$. When $y_i$ (resp.\ the entries of $\bm{s}_i$) is categorical with $N$ categories, at most $N-1$ linearly independent directions are needed. This gives an upper bound for $d$ (resp.\ $m$). In both the categorical and continuous cases, one can use the methods proposed by \cite{Li91} and \cite{Schott94}. For example, \cite{Li91} introduced an eigenvalue-based sequential test for the true SDR subspace dimension $d_\star$. Based on the test, \cite{Tan18} provided a lower bound for the eigenvalue $\tau_{d_\star}$. We use the lower bound to select the SDR dimension $d$. In particular, we choose the largest dimension $d$ such that $\tau_d$ satisfies the lower bound.

\paragraph{Estimating the Fair Subspace}
Given $\bm{W}$, we invoke \cref{prop:SDR-fairness} to compute the fair subspace $\mathcal{F}$ which is described by $\bm{Q}$. In the RKHS setting, the covariance matrix $\Var\left(X\right)$ in \cref{prop:SDR-fairness} is replaced by the covariance operator $\Var\left(\kappa\left(\cdot,X\right)\right)$ on $\mathcal{H}_{\kappa}$. Let $\otimes$ denote the tensor product and let $\mu_X \coloneqq \mathbb{E}_X \left[\kappa\left(\cdot,X\right)\right]$, the empirical estimator of $\Var\left(\kappa\left(\cdot,X\right)\right)$, be written
$\mathbb{E}_X \left[\left(\kappa\left(\cdot,X\right) - \mu_X\right) \otimes \left(\kappa\left(\cdot,X\right) - \mu_X\right)\right] \approx \frac{1}{n} \bm{\phi} \bm{\Gamma}_n \otimes \bm{\phi} \bm{\Gamma}_n$.
\cref{prop:SDR-fairness} states that for all $i \in \left[m\right]$ and $j \in \left[r\right]$, $\bm{Q}$ satisfies
    $\left<\bm{\phi}\bm{W}_i, \left(\frac{1}{n} \bm{\phi} \bm{\Gamma}_n \otimes \bm{\phi} \bm{\Gamma}_n\right) \bm{\phi}\bm{Q}_j\right>
    = \frac{1}{n}\bm{W}_i^\top \bm{K} \bm{\Gamma}_n \bm{K} \bm{Q}_j
    = \bm{0}.$
Thus, the columns of $\bm{Q}$ are given by a basis of the nullspace of $\bm{W}^\top \bm{K} \bm{\Gamma}_n \bm{K}$.

A subtlety in estimating the fair subspace for EO and EOP, as described in \cref{sec:indep}, is that only a subset of the training data with certain value of $Y$ is used. In this case, $\bm{W}$ and $\bm{Q}$ both will have a reduced number of rows. This is not an issue as the fair subspace is still a subspace of $\mathcal{H}_{\kappa,n}$, and the pseudo-code in \cref{sec:code} shows how to handle this case.

\subsection{Controlling the Trade-off between Accuracy and Fairness}
\label{sec:tradeoff}
We now describe a fairness-accuracy trade-off specified by the maximum principal angle between two subspaces of $\mathcal{H}_{\kappa,n}$. Recall that the $i$-th principal angle $\theta_i$ between $\mathcal{F}$ and $\mathcal{G}$ is defined as \citep{Stewart90,Golub13}:
\begin{align*}
    \cos \theta_i \coloneqq \max_{\substack{f_i\in\mathcal{F}, \left\|f_i\right\| \leq 1\\ \forall j<i: \left<f_i,f_j\right> = 0}} \max_{\substack{g_i\in\mathcal{G}, \left\|g_i\right\| \leq 1\\ \forall j<i: \left<g_i,g_j\right> = 0}} \left<f_i,g_i\right>.
\end{align*}
If the largest principal angle $\max_i \theta_i$ equals $0$, $\mathcal{F}$ and $\mathcal{G}$ coincide. Based on this idea, we consider constructing the hypothesis class of the model as a subspace $\mathcal{M}$ of $\mathcal{H}_{\kappa,n}$ such that the largest principal angle between $\mathcal{M}$ and $\mathcal{F}$ is small. Intuitively, functions in $\mathcal{M}$ would then be approximately fair.

More formally, our goal is to enforce the distance between $\mathcal{M}$ and $\mathcal{F}$ as measured by the largest principal angle to be no greater than a given threshold. This is equivalent to requiring the cosine of the largest principal angle to be no less than a parameter $0 \leq \epsilon \leq 1$ specified by the user. Recall that the cosines of principal angles are the singular values of the projection of an orthonormal basis of one subspace onto an orthonormal basis of the other \citep{Golub13}. A direct method for finding an $\mathcal{M}$ that satisfies the principal angle constraint is by reversing the well-known Wedin's bound for the perturbation of singular subspaces \citep{Wedin72}. However, a limitation that inherits from the bound is the dependency on the eigengap. Therefore, we instead consider a simple construction of $\mathcal{M}$ given by:
\begin{align*}
    \mathcal{M} \coloneqq \Span\left\{a_i e_i + b_i u_i\right\}_{i=1}^d
\end{align*}
for some orthonormal set of functions $\left\{e_i\right\}_{i=1}^r$ in $\mathcal{F}$ and orthonormal set of functions $\left\{u_j\right\}_{j=1}^d$ in $\mathcal{G}$. With careful choices of $a_i$, $b_j$, as well as the orthonormal sets, we show that the above hypothesis class satisfies the principal angle constraint as well as several desirable properties. 

First, we compute an orthonormal basis for $\mathcal{F}$ and $\mathcal{G}$ by performing the eigenvalue decompositions
$\bm{Q}^\top\bm{K}\bm{Q}\bm{M} = \bm{M}\bm{\Lambda}$ and $\bm{A}^\top\bm{K}\bm{A}\bm{T} = \bm{T}\bm{\Omega}$. The columns of $\bm{M}$ and $\bm{T}$ are eigenvectors, while $\bm{\Lambda}$ and $\bm{\Omega}$ are diagonal containing the corresponding eigenvalues, i.e., $\lambda_i = \Lambda_{ii}$ and $\omega_i \coloneqq \Omega_{ii}$. It is easy to see that $\mathcal{F}$ and $\mathcal{G}$ have the following orthonormal bases:
\begin{align}
\label{eq:ortho-bases}
    \mathcal{F} \coloneqq \Span\left\{\lambda_i^{-1/2}\bm{\phi}\bm{Q}\bm{M}_i\right\}_{i=1}^r, \quad
    \mathcal{G} \coloneqq \Span\left\{\omega_i^{-1/2}\bm{\phi}\bm{A}\bm{T}_i\right\}_{i=1}^d.
\end{align}

Using the orthonormal bases \eqref{eq:ortho-bases}, \cref{thm:proj-pert} gives the hypothesis space $\mathcal{M}$ for the model which is bounded between the fair RKHS $\mathcal{F}$ and the predictive RKHS $\mathcal{G}$ through $\epsilon$ specifying the cosine of the largest principal angle between $\mathcal{M}$ and $\mathcal{F}$.
\begin{theorem}
\label{thm:proj-pert}
Let $\bm{\Lambda}^{-1/2}\bm{M}^\top\bm{Q}^\top\bm{K}\bm{A}\bm{T}\bm{\Omega}^{-1/2} = \bm{U}\bm{\Sigma}\bm{V}^\top$ be the thin singular value decomposition with singular values $\sigma_i \coloneqq \Sigma_{ii}$, and let the hypothesis class of the model be
\begin{gather}
\begin{split}
     \mathcal{M} = \Span\left\{
     \bm{\phi} \left[
     \left(\gamma_i - \rho_i \sigma_i\right) \bm{Q}\bm{M}\bm{\Lambda}^{-1/2}\bm{U}_i +
     \rho_i \bm{A}\bm{T}\bm{\Omega}^{-1/2}\bm{V}_i
     \right]
     \right\}_{i=1}^d \label{eq:model-basis}
\end{split}
\end{gather}
with
     $\gamma_i \coloneqq \max\left\{\sigma_i,\epsilon\right\}$,
     and $\rho_i \coloneqq \sqrt{\frac{1-\gamma_i^2}{1-\sigma_i^2}}$ if $\sigma_i < 1$ and $\rho_i \coloneqq 0$ if $\sigma_i = 1$ for $i=1,2,\cdots,d$.
Denote by $\sigma_{\text{min}} \coloneqq \min_i \sigma_i$ and let $\mathcal{P}_{\mathcal{F}}$, $\mathcal{P}_{\mathcal{G}}$, and $\mathcal{P}_{\mathcal{M}}$ be the orthogonal projection operators onto the fair RKHS $\mathcal{F}$, predictive RKHS $\mathcal{G}$, and the model RKHS $\mathcal{M}$, respectively. Then, the following operator norms hold:
\begin{align}
    \left\|\mathcal{P}_{\mathcal{F}} - \mathcal{P}_{\mathcal{M}}\right\|
    &= \sqrt{1 - \max\left\{\epsilon^2,\sigma_{\text{min}}^2\right\}} \label{eq:fair-proj-pert}\\
    \left\|\mathcal{P}_{\mathcal{G}} - \mathcal{P}_{\mathcal{M}}\right\|
    &= \max\left\{0, \epsilon\sqrt{1-\sigma_{\text{min}}^2} - \sigma_{\text{min}}\sqrt{1-\epsilon^2}\right\}.
    \label{eq:accu-proj-pert}
\end{align}
\end{theorem}

For the case where $\epsilon = 1$, the basis of \eqref{eq:model-basis} are linear combinations of the orthonormal basis of $\mathcal{F}$ in \eqref{eq:ortho-bases}, and hence $\mathcal{M} \subset \mathcal{F}$. Additionally, it can be verified that $\left\|\mathcal{P}_{\mathcal{F}} - \mathcal{P}_{\mathcal{M}}\right\| = 0$ from \eqref{eq:fair-proj-pert}, and \eqref{eq:accu-proj-pert} becomes $\left\|\mathcal{P}_{\mathcal{G}} - \mathcal{P}_{\mathcal{M}}\right\| = \sqrt{1-\sigma_{\text{min}}^2}$ which is the sine of the largest principal angle between $\mathcal{F}$ and $\mathcal{G}$ as desired. Similarly, if we set $\epsilon=0$ we get $\mathcal{M} = \mathcal{G}$.

The key utility of \cref{thm:proj-pert} involves bounding the difference between the model obtained using $\mathcal{M}$ and an optimal fair (or predictive) model. In particular,
\cref{eq:generalization} shows that $\mathcal{M}$ contains a function which approximates the optimal fair model $f_{\text{fair}} \in \mathcal{H}_\kappa$. Let $\delta_{\bm{x}} f \coloneqq f\left(\bm{x}\right)$ be the evaluation functional.
Then, for any $\bm{x} \in \mathcal{X}$:
\begin{align}
\label{eq:generalization}
\begin{split}
     &\left| \left(\mathcal{P}_{\mathcal{M}} f_{\text{fair}}\right)\left(\bm{x}\right) - f_{\text{fair}}\left(\bm{x}\right)\right|\\
     &= \left\| \delta_{\bm{x}}\left(\mathcal{P}_{\mathcal{M}} f_{\text{fair}}\right) - \delta_{\bm{x}} f_{\text{fair}}\right\|\\
     &\leq \left\|\delta_{\bm{x}}\right\| \left\| \mathcal{P}_{\mathcal{M}} f_{\text{fair}} - f_{\text{fair}}\right\|\\
     &= \left\|\delta_{\bm{x}}\right\| \left\| \mathcal{P}_{\mathcal{M}} f_{\text{fair}} - \mathcal{P}_{\mathcal{F}} f_{\text{fair}} + \mathcal{P}_{\mathcal{F}} f_{\text{fair}} - f_{\text{fair}}\right\|\\
     &\leq \left\|\delta_{\bm{x}}\right\| \left(
     \left\| \mathcal{P}_{\mathcal{M}} - \mathcal{P}_{\mathcal{F}} \right\| \left\|f_{\text{fair}}\right\| + \left\| \mathcal{P}_{\mathcal{F}} f_{\text{fair}} - f_{\text{fair}} \right\|
     \right).
\end{split}
\end{align}
Here, $\left\|\delta_{\bm{x}}\right\|$ and $\left\|f_{\text{fair}}\right\|$ are bounded from the property of the RKHS, and the last norm in \eqref{eq:generalization} converges in probability to zero at rate $O_P\left(n^{-1/4}\right)$ by the consistency of $\mathcal{F}$ and $\mathcal{G}$ \citep{Wu13}. Together with \eqref{eq:fair-proj-pert}, \eqref{eq:generalization} sheds light on the impact of $\epsilon$ on the generalization of the fairness criteria; similar arguments can be made for an optimal predictive model.

\section{Application to Fair GPs}
\label{sec:fairgp}

In this section, we demonstrate the utility of our approach by constructing a fair Gaussian process (FGP) which can be used to develop a class of fair models under the Bayesian framework \citep{Rasmussen06}. In particular, the FGP specifies a prior over functions in $\mathcal{M}$ that satisfy the fairness criteria discussed in \cref{sec:indep}.

Recall that a GP $\left\{f\left(x\right): x \in \mathcal{X}\right\}$ is specified by a mean function and a covariance function. The covariance function characterizes the class of functions, i.e., the curves of $f\left(\cdot\right)$, the GP can realize. The FGP is a GP equipped with a covariance function that ensures that any sample path $f\left(x\right)$ is fair. To obtain the covariance function, we use the integral representation of GPs \citep{Ito54}:
\begin{align}
\label{eq:igp}
    f\left(\bm{x}\right) = \int_{\mathcal{X}} \kappa\left(\bm{x},\bm{z}\right) \nu\left(\bm{z}\right) d \mu\left(\bm{z}\right),
\end{align}
where $\nu: \mathcal{X} \times \Omega \mapsto \mathbb{R}$ is another GP on a probability space $\left(\Omega,\mathcal{F},P\right)$ and $\mu$ is the measure on $\mathcal{X}$. Clearly, $f\left(\cdot\right)$ given by \eqref{eq:igp} is a GP. Without loss of generality, we assume the GP has mean zero, then the covariance function $\Cov\left(f\left(\bm{x}\right), f\left(\bm{z}\right)\right)$ is written 
\begin{align*}
     \int_{\mathcal{X}} \int_{\mathcal{X}} \kappa_\nu\left(\bm{s},\bm{t}\right) \kappa\left(\bm{x},\bm{s}\right) \kappa\left(\bm{z},\bm{t}\right) d \mu\left(\bm{s}\right) d \mu\left(\bm{t}\right),
\end{align*}
where $\kappa_\nu\left(\cdot,\cdot\right)$ is the covariance function of the GP $\nu\left(\cdot\right)$. A key property of \eqref{eq:igp} we use to construct the FGP is that functions in the form of \eqref{eq:igp} are contained in the RKHS generated by kernel $\kappa\left(\cdot,\cdot\right)$ \citep{Pillai07,Tan18}. By replacing $\kappa\left(\cdot,\cdot\right)$ in \eqref{eq:igp} with the reproducing kernel of $\mathcal{M}$, we obtain the desired FGP which inherits the fairness as well as accuracy guarantees of $\mathcal{M}$. It is worth noting that the representation $\mathcal{M}$ in \eqref{eq:model-basis} can be computed independent of data. This can be done using a likelihood for SDR subspaces \citep{Cook09}.

In practice, we use a sample version of the FGP for improved computational efficiency. Consider the sample average of \eqref{eq:igp} given by
\begin{align}
\label{eq:sample-igp}
    f_n\left(\bm{x}\right) \coloneqq \frac{1}{n} \sum_{i=1}^n \nu\left(\bm{x}_i\right) \kappa_{\mathcal{M}}\left(\bm{x},\bm{x}_i\right),
\end{align}
which converges in probability to \eqref{eq:igp} at rate $O_p\left(n^{-1/2}\right)$ by the central limit theorem for Hilbert spaces \citep{Ledoux91}. The covariance function of the sample FGP has two parameters, namely the covariance function $\kappa_\nu\left(\cdot,\cdot\right)$ of $\nu\left(\cdot\right)$ and the reproducing kernel $\kappa_{\mathcal{M}}\left(\cdot,\cdot\right)$ of the RKHS $\mathcal{M}$. 

Finally, we give a reparameterization to simplify the sample FGP. Let $\bm{\phi}\bm{E}_i$ represent the $i$-th basis function of \eqref{eq:model-basis}, and denote by $\bm{\Pi}\left(\bm{z}\right) \coloneqq \left(\kappa\left(\bm{z},\bm{X}\right) - \bm{1}\bm{1}_n^\top\bm{K}/n\right)\bm{E}$, where $\bm{K}$ is the kernel matrix of $\kappa$ as defined in \eqref{eq:rkhs-eig}. The sample FGP can be rewritten as
\begin{align}
\label{eq:FGP}
f_n\left(\cdot\right) \sim \mathcal{G P}\left(\bm{0},
\bm{\Pi}\left(\cdot\right)\bm{\Lambda} \bm{\Pi}\left(\cdot\right)^\top\right),
\end{align}
where the covariance function has only a single hyperparameter, a positive definite matrix $\bm{\Lambda}$. Now, it is straightforward to choose  $\bm{\Lambda}$ as to maximizes the marginal likelihood while training the FGP \eqref{eq:FGP}.

\section{Experiments}
\label{sec:exp}

We present experiments on five real datasets to: \emph{(1)} demonstrate the efficacy of our approach in mitigating discrimination while maintaining prediction accuracy; \emph{(2)} characterize the empirical behavior of the algorithms developed in \cref{sec:alg}; and \emph{(3)} highlight the ability of our method to handle multiple, possibly continuous, protected attributes.

We adapt the experimental setup, including the processed datasets, code, as well as configurations, used in prior work~\citep{Donini18,komiyama2018nonconvex} to compare the proposed FGP\footnote{The datasets and our Matlab implementation of the FGP are available at \url{https://github.com/ZilongTan/fgp}.} against several approaches: a standard GP trained on an adversarially-fair representation~\citep{madras2018learning} (LAFTR-GP), fairness-constrained ERM~\citep{Donini18}, both the linear (Linear-FERM) and nonlinear (FERM) variants, and non-convex fair regression \citep{komiyama2018nonconvex} (NCFR) which supports settings with multiple protected attributes. We also report the results of a standard GP with no fairness objective.

We measure fairness conditions empirically using the absolute correlation coefficient, as it can be generalized to the regression setting. Specifically, we compute the population $\left|\Corr\left(\Yhat,S\right)\right|$ as the SP risk score, $\left|\Corr\left(\Yhat,S\right)\right|$ on individuals with $Y=1$ for EOP, and EO is given by the maximum absolute correlation on individuals with $Y=1$ and $Y=0$. All scores are calculated on holdout test data. For the experiments, we do not consider proxy non-discrimination as it relies on certain model structures, and is not comparable to our chosen baselines. Finally, for the baseline methods we use the code published online by their respective authors, and for the GPs we use a linear mean and a radial basis covariance.

\subsection{Fair Classification}

\begin{figure*}[t]
\centering
\begin{adjustbox}{width=\textwidth}
\input{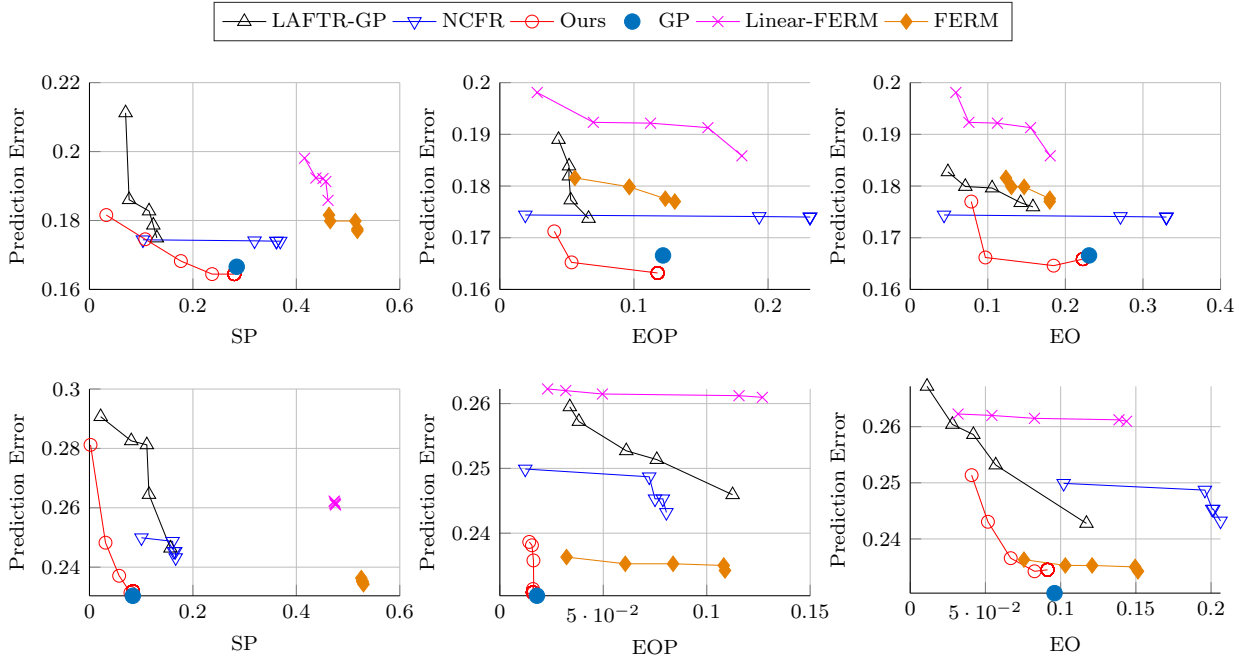}
\end{adjustbox}
\caption{Comparing the accuracy-fairness trade-offs on the Adult (first row) and Compas (second row) datasets. The prediction error denotes the misclassification rate.}
\label{fig:classification}
\end{figure*}

\begin{figure}[t]
\centering
\begin{adjustwidth*}{}{-3.6em}
\input{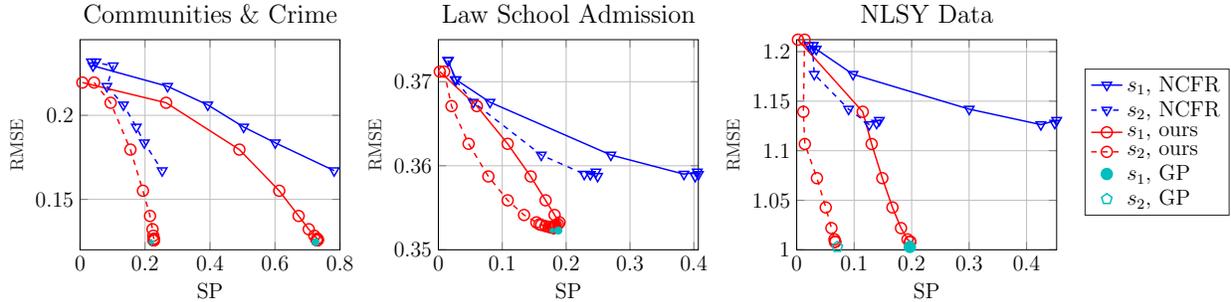}
\end{adjustwidth*}
\caption{Regression results with two protected attributes $s_1$ and $s_2$.}
\label{fig:regression}
\end{figure} 

A primary goal of our approach is to enforce a specified fairness criterion with minimal loss in accuracy. We evaluate how each of the fairness conditions are satisfied empirically on two standard datasets: the Adult income dataset~\citep{Lichman13}, and the Compas recidivism risk score data~\citep{Angwin13}. We illustrate how the fairness score and prediction error react to various choices of the fairness-accuracy tradeoff $\epsilon$. For both datasets, we use gender as the single protected attribute.

\cref{fig:classification} compares different methods for achieving each fairness goal (column). First, observe that Linear-FERM and FERM do not meet SP on both datasets. This is because Linear-FERM and FERM only target EOP. Also note that the accuracy of our approach tends to converge to that of standard GP, which is expected as setting the trade-off $\epsilon = 0$ in \eqref{eq:model-basis} yields this model. Some baselines attain the the best fairness, e.g., LAFTR-GP delivers the lowest EO on Compas, at the cost of accuracy. However, overall our approach generally achieves greater accuracy for a given level of fairness.

\subsection{Fair Regression with Multiple Protected Attributes}

We consider a regression setting with two protected attributes. For this evaluation, we use three real datasets with continuous target values, namely the UCI Communities and Crime~\citep{redmond2002data}, the National Longitudinal Survey of Youth (NLSY)~\citep{nlsy}
and the Law School Admissions Council~\citep{lsac} datasets.
The protected attribute pairs $\left(s_1,s_2\right)$ used for these datasets are respectively ({\it race}, {\it origin}), ({\it gender}, {\it age}), and ({\it race}, {\it age}).

Note that NCFR is the only baseline that handles multiple protected attributes. In addition, EO and EOP are defined in the context of binary classification, so they are not suitable in this regression experiment. We use the root mean squared error (RMSE) to measure the prediction error.

\cref{fig:regression} depicts the prediction error as well as SP for each protected attribute. As stricter fairness conditions are enforced, the RMSE climbs. Across these datasets, our approach achieves consistently improved accuracy. Interestingly, the curves correspond to our approach are generally steeper than the curves of NCFR, suggesting more effective fairness-accuracy trade-offs.

\section{Conclusions}
\label{sec:concl}
We have presented a novel and theoretically principled method for learning fair representations for kernel models, which also enables users to systemically navigate the fairness-accuracy trade-off.  We apply our approach to obtain a fair Gaussian process, demonstrating competitive empirical performance on several datasets relative to state-of-the-art methods. Our work hinges on the idea of learning a model-aware representation, along with the key insight that several popular fairness notations can be reformulated as sufficient dimension reduction (SDR) problems. Future work involves supporting additional fairness notions like calibration and accuracy parity through additional model assumptions, developing more scalable algorithms using randomized approximations, and generalizing the strategy of learning model-aware representations to other model classes.

\section*{Acknowledgements}

This work was supported in part by DARPA FA875017C0141, the National Science Foundation grants IIS1705121, IIS1838017, CNS1704845, an Okawa Grant, a Google Faculty Award, an Amazon Web Services Award, a JP Morgan A.I. Research Faculty Award, and a Carnegie Bosch Institute Research Award. Any opinions, findings and conclusions or recommendations expressed in this material are those of the author(s) and do not necessarily reflect the views of DARPA, the National Science Foundation, or any other funding agency.

\appendix
\section{Appendix}

\subsection{Proofs}
\label{sec:proofs}

\begin{proof}[Proof of \cref{prop:SDR-fairness}]
First, we use a property of elliptically contoured distributions \citep[Corollary 5]{Cambanis81} to obtain
\begin{align*}
	\E\left[\bm{C}^\top X\mid \bm{B}^\top X\right] 
	&= \bm{a} + \Cov\left(\bm{C}^\top X,\bm{B}^\top X\right)\Var^{-1}\left(\bm{B}^\top X\right)\left[\bm{B}^\top X - \E \left(\bm{B}^\top X\right)\right]\\
	&= \bm{a} + \bm{C}^\top \Var\left(X\right) \bm{B} \left(\bm{B}^\top \Var\left(X\right) \bm{B}\right)^{-1} \bm{B}^\top\left(X - \E X\right)
\end{align*}
for some constant $\bm{a}$. From condition \eqref{eq:SDR-S} and the law of total covariance, \begin{align*}
    \Cov\left(S,\bm{C}^\top X\right) 
    &= \Cov\left(f_S\left(\bm{B}^\top X, \epsilon_S\right),\bm{C}^\top X\right) \\
    &= \mathbb{E} \left[\Cov\left(f_S\left(\bm{B}^\top X, \epsilon_S\right),\bm{C}^\top X\mid \bm{B}^\top X, \epsilon_S\right)\right] +\\
    &\phantom{=}\Cov \left[\mathbb{E}\left(f_S\left(\bm{B}^\top X, \epsilon_S\right)\mid \bm{B}^\top X, \epsilon_S\right),\mathbb{E}\left(\bm{C}^\top X\mid \bm{B}^\top X, \epsilon_S\right)\right]\\
    &= \Cov\left(f_S\left(\bm{B}^\top X, \epsilon_S\right), \mathbb{E}\left(\bm{C}^\top X\mid \bm{B}^\top X\right)\right)\\
    &= \Cov\left(f_S\left(\bm{B}^\top X, \epsilon_S\right),X\right)
    \bm{B} \left(\bm{B}^\top \Var\left(X\right) \bm{B}\right)^{-1} \bm{B}^\top \Var\left(X\right) \bm{C}.
\end{align*}
Thus, we have $\Cov\left(S,\bm{C}^\top X\right) = \bm{0}$ if $\bm{B}^\top \Var\left(X\right) \bm{C} = \bm{0}$ which implies that the columns of $\bm{C}$ lie in the nullspace of $\Var\left(X\right) \bm{B}$.
\end{proof}

\begin{proof}[Proof of \cref{thm:proj-pert}]
We first show that the basis \eqref{eq:model-basis} of the classifier hypothesis RKHS is orthonormal, and then compute the canonical angles using the basis \eqref{eq:model-basis} and an orthonormal basis of $\mathcal{F}$. Denote by $\xi_i \coloneqq \left(\gamma_i - \rho_i \sigma_i\right) \bm{Q}\bm{M}\bm{\Lambda}^{-1/2}\bm{U}_i + \rho_i \bm{A}\bm{T}\bm{\Omega}^{-1/2}\bm{V}_i$ the $i$-th basis function in \eqref{eq:model-basis}, we have
\begin{align*}
    \left<\xi_i, \xi_j\right>
     &= \left(\gamma_i - \rho_i \sigma_i\right)\left(\gamma_j - \rho_j \sigma_j\right) \bm{U}_i^\top\bm{U}_j
     + \rho_i \rho_j \bm{V}_i^\top\bm{V}_j + 2\rho_i\sigma_i\left(\gamma_i - \rho_i \sigma_i\right) \mathbbm{1}_{i=j}\\
     &= \left[\left(\gamma_i - \rho_i \sigma_i\right)^2 + \rho_i^2 + 2\rho_i\sigma_i\left(\gamma_i - \rho_i \sigma_i\right)\right] \mathbbm{1}_{i=j}\\
     &= \mathbbm{1}_{i=j},
\end{align*}
where the first equality follows from the orthonormal basis \eqref{eq:ortho-bases}. This shows that \eqref{eq:model-basis} is an orthonormal basis. Using the orthonormal basis $\left\{\psi_i \coloneqq \bm{\phi}\bm{Q}\bm{M}\bm{\Lambda}_i^{-1/2}\right\}_{i=1}^r$ of $\mathcal{F}$, we can use the SVD to compute the canonical angles (see e.g., Algorithm 6.4.3 in \citep{Golub13}) as
\begin{align}
    \left[\xi_1,\cdots,\xi_d\right]^\top \left[\psi_1,\cdots,\psi_r\right] 
    = \text{diag}\left(\gamma_i - \rho_i \sigma_i\right) \bm{U}^\top
     + \text{diag}\left(\rho_i\right) \bm{\Sigma} \bm{U}^\top
    = \bm{I}_d \text{diag}\left(\gamma_i\right) \bm{U}^\top. \label{eq:svd-angle}
\end{align}
Here, $\text{diag}\left(d_i\right)$ denotes the diagonal matrix with diagonal elements $d_i$. Note that the last term in \eqref{eq:svd-angle} is the (thin) SVD, and the singular values $\gamma_i$ are the canonical angles between $\mathcal{M}$ and $\mathcal{F}$. Finally, we relate the canonical angles to the operator norm in \eqref{eq:fair-proj-pert}. Recall that the orthogonal projector can be expressed as the tensor product $\mathcal{P}_{\mathcal{F}} = \sum_{i=1}^r \psi_i \otimes \psi_i$, and $\mathcal{P}_{\mathcal{F}} h = \sum_{i=1}^r \left<h,\psi_i\right> \psi_i$. We have
\begin{align*}
    \left\|\mathcal{P}_{\mathcal{F}} - \mathcal{P}_{\mathcal{M}}\right\| 
    &= \left\|\left(\mathcal{P}_{\mathcal{F}} + \mathcal{P}_{\mathcal{F}}\right)\left(\mathcal{P}_{\mathcal{F}} - \mathcal{P}_{\mathcal{M}}\right)\left(\mathcal{P}_{\mathcal{M}} + \mathcal{P}_{\mathcal{M}}\right)\right\|\\
    &= \left\|\mathcal{P}_{\mathcal{F}}\mathcal{P}_{\mathcal{M}} - \mathcal{P}_{\mathcal{F}}\mathcal{P}_{\mathcal{M}}\right\|\\
    &= \sup_{h\in\mathcal{H}_{\kappa,n}\colon\left\|h\right\| \leq 1}
    \left(\left\|\mathcal{P}_{\mathcal{F}}\mathcal{P}_{\mathcal{M}} h \right\| + \left\|\mathcal{P}_{\mathcal{F}}\mathcal{P}_{\mathcal{M}} h\right\|\right),
\end{align*}
where $\mathcal{F}$ and $\mathcal{M}$ represent respectively the orthogonal complements of $\mathcal{F}$ and $\mathcal{M}$. It can be shown that $\mathcal{P}_{\mathcal{F}}\mathcal{P}_{\mathcal{M}}$ and $\mathcal{P}_{\mathcal{F}}\mathcal{P}_{\mathcal{M}}$ have the same nonzero singular values which are the sines of the principal angles between $\mathcal{F}$ and $\mathcal{M}$ (see e.g., p.249 of \citealp{Stewart01}). From \eqref{eq:svd-angle}, these principal angles are $\arccos\left(\gamma_i\right)$. Thus, we obtain $\left\|\mathcal{P}_{\mathcal{F}} - \mathcal{P}_{\mathcal{M}}\right\| = \sqrt{1-\min_i\gamma_i^2}$. To obtain \eqref{eq:accu-proj-pert}, one can simply apply the trigonometric identity of sines yielding 
\begin{align*}
    \left\|\mathcal{P}_{\mathcal{G}} - \mathcal{P}_{\mathcal{M}}\right\|
    = \gamma_{\text{min}} \sqrt{1 - \sigma_{\text{min}}^2} - \sigma_{\text{min}}\sqrt{1-\gamma_{\text{min}}^2}
    = \max\left\{0, \epsilon\sqrt{1-\sigma_{\text{min}}^2} - \sigma_{\text{min}}\sqrt{1-\epsilon^2}\right\},
\end{align*}
where we denote by $\gamma_{\text{min}} \coloneqq \min_i \gamma_i$.
\end{proof}

\subsection{Implementation}
\label{sec:code}

\cref{alg:learn} gives the Matlab-style pseudo-code for our approach which can handle multiple protected attributes. This algorithm use the SDR procedure described in \cref{alg:sdr} to compute the desired model representation with a specified trade-off $\epsilon$.

\begin{figure}[ht]
  \centering
  \begin{minipage}{\linewidth}
\begin{algorithm}[H]
\SetAlgoLined
\SetNlSty{texttt}{[}{]}
   \nl Initialize $\bm{W} = \left[\;\right]$, $n$ with the number of rows of $\bm{K}$, as well as indices $\texttt{pos} = \left(\bm{y}==1\right)$ and $\texttt{neg} = \left(\bm{y}\neq 1\right)$.
   
   \nl \ForEach{column $\bm{s}$ of $\bm{S}$}{
   \nl \eIf{EqualizedOdds {\bf or} EqualityOfOpportunity}{
   
   \nl Set $\bm{B} = \bm{0}_{n\times m}$ and update $\bm{B}\left(\texttt{pos}:\right) = \texttt{SDR}\left(\bm{K}\left(\texttt{pos},\texttt{pos}\right),\bm{s},m\right)$.
   
   \nl Append basis $\bm{B}$ to $\bm{W}$: $\bm{W} = \left[\bm{W} \enspace \bm{B}\right]$.
   
   \nl\If{EqualizedOdds}{
   \nl Set $\bm{B} = \bm{0}_{n\times m}$, then $\bm{B}\left(\texttt{neg},:\right) = \texttt{SDR}\left(\bm{K}\left(\texttt{neg},\texttt{neg}\right),\bm{s},m\right)$.
   
   \nl Let $\bm{W} = \left[\bm{W} \enspace \bm{B}\right]$.
   } }{
   \nl Compute $\bm{B} = \texttt{SDR}\left(\bm{K},\bm{s},m\right)$, and update $\bm{W} = \left[\bm{W} \enspace \bm{B}\right]$.
  }
  }
  
  \nl {\bf Predictive Subspace:} Compute the predictive subspace as the SDR subspace $\bm{A} = \texttt{SDR}\left(\bm{K},\bm{y},d\right)$.
  
  \nl {\bf Fair Subspace:} Obtain $\bm{K}^\prime$ by subtracting the mean of each column of $\bm{K}$. Let $\tilde{\bm{K}} = \bm{K}^{\prime\top}\bm{K}^\prime$, and use QR decomposition to compute $\bm{Q}$ as the nullspace basis of the column space of $\tilde{\bm{K}}\bm{W}$.
  
  \nl Perform the eigenvalue decompositions to obtain \cref{eq:ortho-bases}, and then use \cref{thm:proj-pert} to compute $\bm{E}$.
  
\caption{$\bm{E} = \texttt{MBasis}\left(\bm{K},\bm{y},\bm{S},m,d,\epsilon\right)$ --- Compute the basis $\bm{\phi}\bm{E}$ for $\mathcal{M}$}
\label{alg:learn}
\end{algorithm}
\end{minipage}
\end{figure}

\begin{figure}[ht]
  \centering
  \begin{minipage}{\linewidth}
\begin{algorithm}[H]
\SetAlgoLined
\SetNlSty{texttt}{[}{]}
   \nl Sort $\bm{s}$ such that $s\left(\texttt{idx}\right)$ is non-decreasing. Let $\texttt{invIdx}$ be the inverse of $\texttt{idx}$ satisfying $\texttt{idx}\left(\texttt{invIdx}\right) = 1:n$, where $n$ is the number of rows of $\bm{K}$.
   
   \nl Slice $\bm{s}$ approximately evenly as described in \cref{sec:learn-basis} such that entries with the same value are in the same partition. Denote by $n_i$ the size of partition $i$.
   
   \nl Initialize $\eta = 10^{-4}$, i.e., a small constant. Let $\bm{K}^\prime \coloneqq \bm{K}\left(\texttt{idx},\texttt{idx}\right)$, and solve $
    \bm{\Gamma}_n\bm{K}^\prime \bm{A}_i = \tau_i \left[\text{diag}\left(\bm{\Gamma}_{n_i}\right)\bm{K}^\prime + n \eta\bm{I}_n\right] \bm{A}_i$ for $\bm{A}$.

   \nl Return $\bm{W} = \bm{A}\left(\texttt{invIdx},:\right)$.
\caption{$\bm{W} = \texttt{SDR}\left(\bm{K},\bm{s},m\right)$ --- Compute the SDR subspace $\bm{\phi}\bm{W}$}
\label{alg:sdr}
\end{algorithm}
\end{minipage}
\end{figure}

\bibliographystyle{plainnat}
\bibliography{biblio}

\end{document}